\documentclass{article}
\usepackage{graphicx}
\usepackage{subcaption}
\usepackage{float}
\usepackage{wrapfig}    %

\usepackage{enumitem}

\usepackage[preprint]{nips/neurips_2025}

\usepackage[utf8]{inputenc} %
\usepackage[T1]{fontenc}    %
\usepackage{hyperref}       %

\usepackage{url}            %
\usepackage{booktabs}       %
\usepackage{amsfonts}       %
\usepackage{nicefrac}       %
\usepackage{microtype}      %
\usepackage{xcolor}         %
\usepackage{tikz}
\usepackage{algorithm}
\usepackage{algorithmic}
\usepackage{amsmath}
\usepackage{amssymb}
\usepackage{mathtools}
\usepackage{amsthm}
\theoremstyle{plain}
\newtheorem{theorem}{Theorem}[section]

\newtheorem{lemma}[theorem]{Lemma}
\newtheorem{corollary}[theorem]{Corollary}
\theoremstyle{definition}
\newtheorem{definition}[theorem]{Definition}

\theoremstyle{remark}

\usepackage{cleveref}

\title{Chordless Structure: A Pathway to Simple and Expressive GNNs}

\usepackage{authblk} 

\author[1]{Hongxu Pan}
\author[2]{Shuxian Hu}
\author[1]{Mo Zhou}
\author[1]{Zhibin Wang}
\author[1]{Rong Gu}
\author[1]{Chen Tian}
\author[1]{Kun Yang}
\author[1]{Sheng Zhong}        

\affil[1]{\footnotesize Nanjing University} 
\affil[2]{\footnotesize Alibaba Group}

\usepackage[compact]{titlesec}
\begin{document}

\maketitle
\renewcommand\Authands{,\ }          
\renewcommand\Authfont{\normalsize} 
\renewcommand\Affilfont{\tiny}      
\begin{abstract}
              Researchers have proposed various methods of incorporating more structured information into the design of Graph Neural Networks (GNNs) to enhance their expressiveness. However, these methods are either computationally expensive or lacking in provable expressiveness. In this paper, we observe that the chords increase the complexity of the graph structure while contributing little useful information in many cases. In contrast, chordless structures are more efficient and effective for representing the graph. Therefore, when leveraging the information of cycles, we choose to omit the chords. Accordingly, we propose a Chordless Structure-based Graph Neural Network (CSGNN) and prove that its expressiveness is strictly more powerful than the k-hop GNN (KPGNN) with polynomial complexity. Experimental results on real-world datasets demonstrate that CSGNN outperforms existing GNNs across various graph tasks while incurring lower computational costs and achieving better performance than the GNNs of 3-WL expressiveness.
       \end{abstract}
    \section{Introduction}
       Graphs have shown their power in representing complex data, from social
       networks\cite{hamilton2017inductive} and chemical molecules\cite{yang2019analyzing}
       to recommendation engines\cite{wu2021self}, by capturing relational
       dynamics. Graph Neural Networks (GNNs), as a powerful tool for graph representation learning, have been widely used in various fields, such as
       social network analysis\cite{qiu2018deepinf}, recommendation systems\cite{ying2018graph},
       and bioinformatics\cite{gainza2020deciphering}. Among the various GNNs, Message
       Passing Neural Networks (MPNNs)~\cite{gilmer2017neural} is the most popular
       family of GNNs, including Graph Convolutional Networks (GCNs)~\cite{kipf2017},
       GraphSAGE~\cite{hamilton2018}, and Graph Attention Networks (GAT)~\cite{velivckovic2017graph}.
       Specifically, MPNNs recursively update node representations based on the
       neighborhood structure, which is similar to the Weisfeiler-Lehman (WL) graph
       isomorphism test\cite{weisfeiler1968reduction} that hashes the
       neighbourhood structure of each node recursively. Therefore, the
       expressiveness of standard MPNNs is limited by the WL test, which indicates
       that MPNNs are unable to distinguish some non-isomorphic graphs.
       The limited expressiveness of MPNNs restricts their performance in
       analyzing complex graph structures, such as substructure counting,
       distinguishing graphs with symmetries and automorphisms, reasoning over
       higher-order interactions, and so on.

       In order to enhance the expressiveness of GNNs, researchers have proposed
       various methods, which can be roughly divided into the following categories.
       \begin{itemize}
              \item \textbf{KPGNN}~\cite{wu2022non,feng2022powerful}: Different from
                     the standard MPNNs considering 1-hop neighborhood, KPGNN
                     utilizes K-hop neighborhood which further includes the
                     cyclic information among k-hop neighbors. However, as
                     indicate by~\cite{wu2022non}, increasing the depth leads to
                     over-smoothing and incurs high computational costs~\cite{chordless2014}, with
                     complexity $\mathcal{O}(n\bar{d}^{k})$ in sparse graphs.

              \item \textbf{k-GNN}~\cite{morris2019weisfeiler}: 
              Observing the MPNNs are bounded by the 1-WL test, researchers imitate the K-WL test to design k-GNNs~\cite{morris2019weisfeiler}. However, the computational and memory complexity of k-GNNs is at least $\mathcal{O}(|V|^{k+1})$ and $\mathcal{O}(|V|^{k})$ respectively, which is infeasible for large graphs~\cite{li2022expressive}.

              \item \textbf{Subgraph GNN}~\cite{sgcgnn}: 
              Recognising that MPNNs are limited to identifying subgraph structures, researchers propose subgraph GNNs to enhance the expressiveness of GNNs by incorporating subgraph information, such as subgraph isomorphism count. However, the subgraph isomorphism count problem is NP-hard and computationally expensive~\cite{mccreesh2018subgraph}. Moreover, \cite{choi2022cycle} indicates that subgraph GNNs are prone to overfitting by injecting too much structural information.
       \end{itemize}

       The existing work highlights the critical role of cycles in enhancing the
       expressiveness of GNNs. However, vast numbers of cycles in the graph suffer
       from over-smoothing and high computational costs.

        \vspace{-6pt}
       \begin{figure}[t]
              \centering
              \begin{subfigure}
                     [b]{.3\linewidth}
                \centering
              \includegraphics[width=\textwidth, trim=8cm 6cm 7cm 4cm, clip]{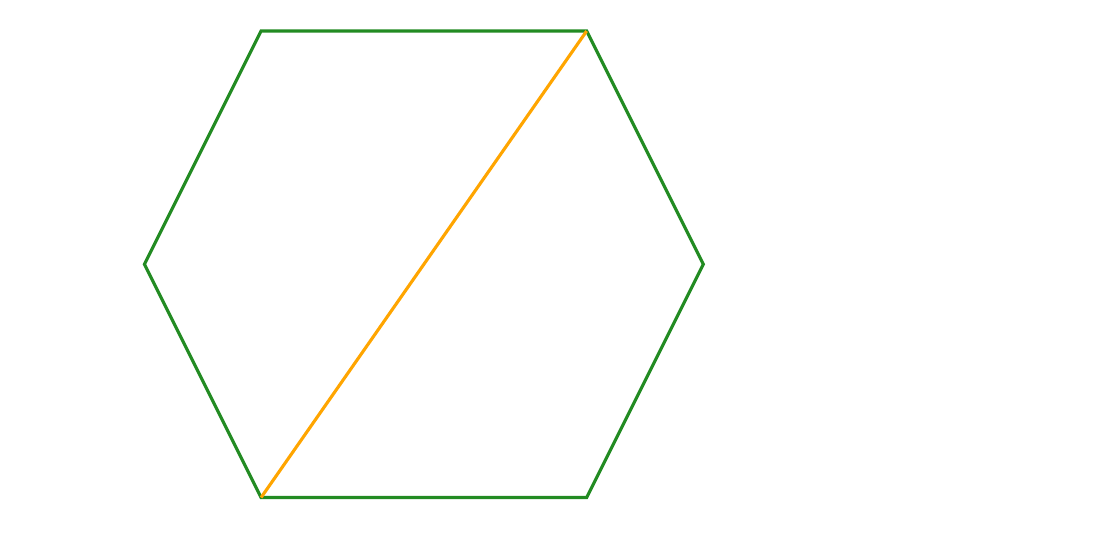}
                     \caption{A chord in a 6-cycle\label{fi:1a} }
              \end{subfigure}
              \hfill
              \begin{subfigure}
                     [b]{.6\linewidth}
                     \small
                     \phantomsection
                     \hspace*{-0.5cm}
                     \begin{tabular}{|lcc|}
                            \hline
                                    & IMDB-B & IMDB-M \\
                            \hline
                           Number of graphs  & 1000   & 1500   \\
                            Average number of nodes &19.8  & 13.0   \\
                            Average number of edges  & 96.5   & 65.9   \\
                            Average number of chordless cycles & 393.1  & 306.2  \\
                            Average number of cycles  & 5641.0 & 5168.6 \\
                            \hline
                     \end{tabular}
                     \caption{Statistics of IMDB-B and IMDB-M datasets}
                     \label{tab:imdb_stats}
              \end{subfigure}
              \vspace{-.1in}
              \caption{Chord and statistics of IMDB datasets.              \vspace{-.1in}
}
              \label{fi:1}
       \end{figure}

       Fortunately, we observe that \emph{if a cycle contains a chord, it can be
       reconstructed by smaller cycles}. Taking money laundering detection as an example, the fund flow chains of money laundering activities often exhibit a circular structure as funds must ultimately revert to specific entities. Meanwhile, fund flow networks often prefer the shortest path: intuitively, criminals tend to transfer funds through the fewest nodes and edges to avoid tracking ~\cite{xu2004fighting}. A cycle with a chord can be broken down into smaller cycles or paths. For example, the chord in the 6-cycle shown in Figure \ref{fi:1}(a) splits it into two 4-cycles, and such smaller cycles better fit the need for rapid fund diversion in money laundering networks.
       
       Specifically, as shown in Figure \ref{fi:1}(a),
       a chord (highlighted in yellow) in a cycle refers to an edge that is not part of the cycle (green edges) but connects two nodes in the cycle. Therefore,
       we propose to use chordless cycles and paths~\cite{satoh2005effective} to
       represent the original, where the rest of the substructures can be reconstructed using the chordless cycles and paths. As shown in Figure \ref{fi:1} (b), the number of chordless cycles (CC) is significantly less than the number of cycles(C)
       while still capturing the necessary structural information, as it enhances the expressiveness of MPNNs by only considering the edge information.

       In this paper, we propose a novel framework for enhancing the expressiveness of Graph Neural Networks by incorporating chordless structures into their
       design\footnote{The code is available at
       \url{https://anonymous.4open.science/r/ul4y4h4/}}. Specifically, we
       compute chordless paths and cycles in the original graphs whose complexity is
       $\mathcal{O}(c(G)(|V|+|E|))$~\cite{chordless2014}. Subsequently, we
       conduct message passing on the chordless paths and cycles to aggregate
       the structural information.
       Our theoretical analysis in Section \cref{sec:4.2} shows that even when restricted to a strictly radius-$k$
       neighborhood, CSGNN achieves strictly greater expressiveness than KPGNN.

       To comprehensively evaluate the effectiveness of CSGNN, we conducted experiments across multiple graph-related tasks, including graph isomorphism testing,
       graph classification, and molecular regression, under consistent experimental settings. Compared to the CY2C model~\cite{choi2022cycle}, which relies on
       extensive cycle-based structural components, CSGNN achieved superior
       performance despite utilizing fewer structural features (see Table \ref{tab:sub1}
       for detailed results).
       Furthermore, we validated the generalizability of chordless structural
       information by augmenting existing popular GNNs with this feature, observing
       consistent performance enhancements across tasks. Remarkably, even when restricted
       to chordless paths, CSGNN matched the accuracy of PathNN~\cite{pathnn}, which
       employs exhaustive path enumeration in its optimal configuration.

    \vspace{-6pt}
\vspace{-6pt}
\section{Preliminaries}

       \subsection{Notation}
       Let $G = (V, E)$ be an undirected graph consisting of a set of nodes $V$
       and a set of edges $E \subseteq V \times V$. We denote by $n$ the number of nodes in $G$ and by $m$ its number of edges. The set $\mathcal{N}(v)$
       represents the neighbors of node $v$, and the set $\mathcal{N}_{k}(v)$ represents the k-hop neighborhood of node $v$. When considering colored graphs $G=(V,E,c)$, we have a color function $c:V\xrightarrow{} \mathcal{C}$, where $\mathcal{C}$ is a finite set of colors. 
       When considering attributed graphs, each node $v \in V$ is endowed with an initial node feature vector denoted by $x
       _{v}$ that can contain categorical or real-valued properties of $v$. And $\langle u,v \rangle$ denotes the edge between node $u$ and node $v$. We denote the natural numbers set $\{1,2,\cdots,n\}:=[n]$.
    
       \subsection{MPNNs and WL Test}
       \begin{figure*}[t]
              \centering
              \includegraphics[
                     width=0.75\textwidth,
                     trim=0cm 7cm 0cm 2cm,
                     clip=true
              ]{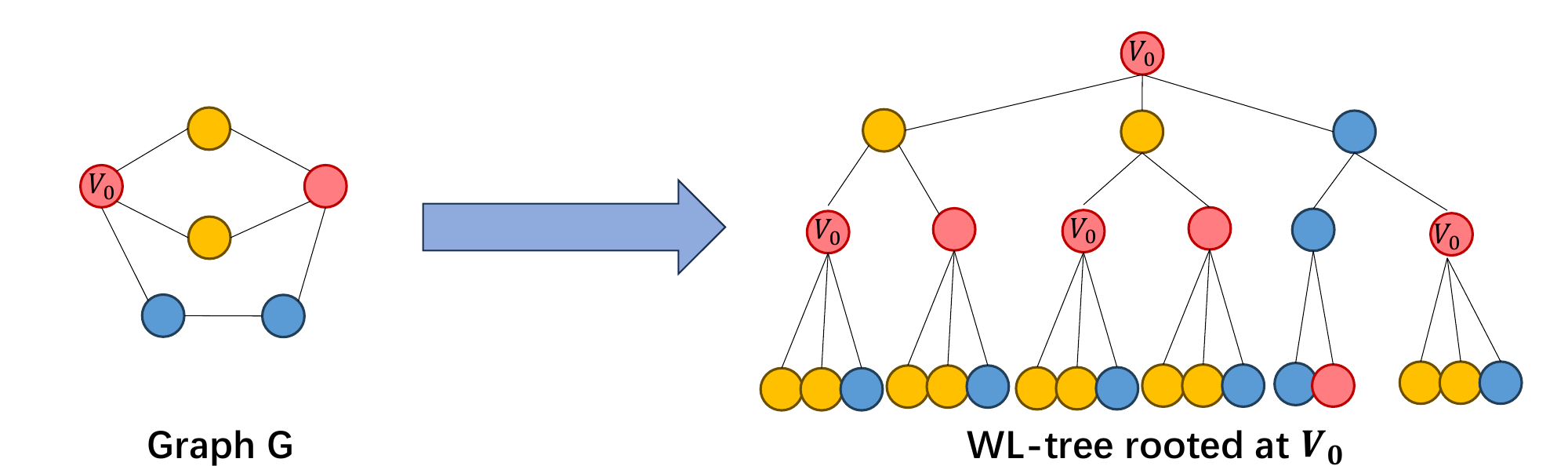}
              \caption{WL-tree of Graph G.}
       \end{figure*}
       \begin{definition}
        Given a coloring algorithm $T$, node $u$ and $v$ are called $T-equivalent$, denoted by $u\sim_{T}v$, if they result in the same color after the termination of the algorithm $T$.  
        \end{definition}
       Currently, most existing GNNs can be designed based on 1-hop message
       passing framework (MPNNs). Given initial node features $h_{v}^{(0)}=x_{v}$,
       the node representation $h_{v}^{(t)}$ at layer $t$ is updated by
       aggregating the features of its neighbors and itself. And finally, the
       whole graph can be represented by the node features.
       \begin{align*}
              h_{v}^{(t+1)} = \mathrm{AGGREGATE}(\{h_{u}^{(t)}: u \in \mathcal{N}(v)\}),
              h_{G} = \mathrm{READOUT}(\{h_{v}^{(k)}: v \in V\}),
       \end{align*}
       where $k$ is the number of iterations in updating processes, and
       $\mathrm{AGGREGATE}$ and $\mathrm{READOUT}$ are the aggregation and
       readout functions. In graph theory, the Weisfeiler-Leman (WL) algorithm\cite{weisfeiler1968reduction} is a graph isomorphism test widely used in graph theory and computer science. It is also as knwon as 1-WL. It uses a hash function to encode the neighborhood structure of each node in a graph, and the whole graph can be represented by the value of each node.
       \begin{align*}
    h_{v} = \mathrm{hash}(\{x_{u} : u \in \mathcal{N}(v)\}),
    h_{G} = \mathrm{READOUT}(\{h_{v} : v \in V\})
\end{align*}

       It is acknowledged that the expressiveness of GNNs is limited by the WL test.
       
       \textbf{KPGNN}\cite{nikolentzos2020k} further uses k-hop neighborhood information to enhance the expressiveness of
       GNNs. The hidden state $h_{v}^{(t)}$ of KPGNN is shown as follows:
       \begin{align*}
    a_{v}^{(t)} = \mathrm{AGGREGATE}(\{h_{u}^{(t-1)} : u \in \mathcal{N}_{k}(v)\}),
    h_{v}^{(t)} = \mathrm{UPDATE}(h_{v}^{(t-1)}, a_{v}^{(t)})
\end{align*}

       where $\mathcal{N}_{k}(v) = \{u |\exists w \in \mathcal{N}_{k-1}(v), u \in \mathcal{N}(v)\} $for $k>1$ and 
       $\mathcal{N}_{1}(v) = \mathcal{N}(v)$ naturally.

       \subsection{High-order GNNs}
       In order to understand the expressiveness of GNNs, we can consider showing
       the expressiveness by comparing it to $KPGNN$, which can be represented
       by $WL-tree$ of a node, which is given by the computational tree of one node
       in the graph like \cite{pathnn}. At each iteration of aggregation, level $l
       +1$ of the WL-tree is constructed by setting the children of any node at the level
       $l$ to its direct neighbors. For a node in the $l$-th layer in $WL-tree$,
       its children in the $(l+1)$-th layer are the neighborhood of the node of it.

       For high-order graph neural networks, Morris et al.~\cite{morris2019weisfeiler} introduced \emph{k-GNNs}, which are based on the \emph{k-dimensional Weisfeiler-Lehman (k-WL) test}. Given a fixed $k$, the method considers all subsets of nodes of size $k$ from the graph. Each such subset is treated as a \emph{super-node}, and two super-nodes are considered neighbors if they differ in exactly one node.

The k-WL procedure iteratively updates a representation (or ``color'') for each super-node by incorporating information from its neighboring super-nodes. To better distinguish between structural and topological aspects of the graph, the neighborhood of each super-node is further divided into \emph{local} and \emph{global} parts. Local neighbors are those where the differing nodes are connected by an edge, while global neighbors are those where the differing nodes are not connected.

\subsection{Subgraphs GNN}
Beyond k-WL-based models, subgraph GNNs like\cite{sgcgnn} offer another approach to enhance expressive power by leveraging the presence and distribution of local substructures in the graph.
\par
As shown in \cite{zhang2023complete}, subgraph GNNs can be formalized as below.

\begin{align*}
    h_G^{(l+1)}(u,v) = \mathrm{AGG}^{(l+1)}\left(\left\{f_i(u,v,G,h_G^{(l)})\right\}_{i=1}^r\right) ,
    f(G) = \mathrm{READ}\left(\left\{h_G^{(L)}(u,v) : u,v \in \mathcal{V}_G \right\}\right),
\end{align*}
where \( h_G^{(l)}(u,v) =h_G^{(0)}(u,v) \) denotes the representation of node \(v\) in the subgraph \(G^u\) centered at node \(u\), at the \(l\)-th layer, \( \tilde{h}_G^u(v) \) is the initial input feature of node \(v\) in subgraph \(G^u\), determined by the graph generation policy (e.g., node marking, distance encoding) and \( f_i \) represents the \(i\)-th atomic operation.

\subsection{Limitation of existing GNNs}
       For standard GNNs, the expressiveness is limited by the WL test, which means
       that GNNs are unable to distinguish some non-isomorphic graphs. For k-hop
       GNNs, \cite{feng2022powerful} has shown that the expressive power of K-hop
       message passing is bounded by 3-WL. For k-GNNs based on k-WL, a key limitation
       of such k-GNNs that achieve WL is their lack of scalability and explainability.
       \cite{zhao2020persistence} shows that these models can only be applied to
       small graphs with small $k$ in real-world tasks. According to \cite{li2022expressive},
       k-WL's memory complexity is at least $\mathcal{O}(n^{k})$ and
       computational complexity is at least $\mathcal{O}(n^{k+1})$ , which is
       infeasible for large graphs. \par

       To solve the problems, we propose a Chordless Structure based Graph Neural Network
       (CSGNN) that uses chordless paths and cycles to enhance the expressiveness
       of GNNs. It comes with more expressiveness beyond WL, but with computational
       efficiency and more explainability.

    \usetikzlibrary{arrows.meta}
\section{Decomposition of Graph into Chordless Cycles and Paths}
       \label{s4}

       \subsection{Chord and Chordless Structure}
       \label{4.1} A path is a simple graph whose nodes can be arranged in a
       linear sequence in such a way that two nodes are adjacent if they are
       consecutive in the sequence, and are nonadjacent otherwise. Likewise, a cycle
       is a simple graph whose nodes can be arranged in a cyclic sequence. Given
       a cycle, if there exists an edge that connects two nodes of the cycle but
       is not part of the cycle, we call it a \emph{chord}. To put it in another
       way, a chord decomposes the original cycle into two smaller cycles,
       conversely, \emph{two smaller cycles can reconstruct the original cycle}.

       This observation motivates us to consider the chordless structure, which is
       enough to encode the whole necessary substructure of the graph.

       Formally, for a graph $G$, a chordless cycle is a cycle that does not contain
       any chord. Similarly, a chordless path is a path that does not contain any
       chord. We respectively denote them by $CC$ and $CP$. In other words, in a
       chordless cycle $CC$, we have
       \[
              \forall u,v \in CC, \quad \langle u, v \rangle \notin CC \rightarrow
              u\notin \mathcal{N}(v).
       \]

       For all chordless cycles/paths of length $k$, we denote them by $CC_{k}$, and We write $u\in CC_k(v)$ if there exists such a chordless cycle containing both $u$ and $v$, and so does $CP_k(v)$ for chordless paths and
       $CP_{k}$, and We write $u\in CC_k(v)$ if there exists such a chordless cycle containing both $u$ and $v$, and so does $CP_k(v)$ for chordless paths.
       \subsection{Decomposition of Graph}
       In this subsection, we will discuss why we choose chordless structures to represent the graph structure.
       \begin{lemma}
              WL can distinguish all non-isomorphic trees.\cite{kiefer2020power}\par
       \end{lemma}
       By constructing the $WL-tree$ of the graph, we can
       get the unique representation of the graph. Since the tree does not contain
       any cycle, we can get a contrapositive proposition of the lemma.
       \begin{corollary}
              \label{thm1} If there exists two graphs $G_1$, $G_2$, $G_1 \sim_{1-WL} G_2$ but $G_1 \ncong G_2$, then there exists a cycle in the graphs.\par
       \end{corollary}
       The corollary \ref{thm1} shows the importance of the cycle in the graph structure.
       Additionally, while the standard GNNs only use the local structure, the
       cycles can be a better choice to represent the graph global structure. In
       graph theory, we have a conjecture.\par \textbf{Conjecture (Cycle Double Cover
       Conjecture)}\cite{seymour1979sums} For any undirected graph $G$, there exists a cycle double
       cover of $G$ if $G$ is bridgeless.\par

       Since the chordless cycle can be considered as a minimal cycle, we can easily find the lemma below.
       \begin{lemma}
              If a node is in a cycle, then the node is in a chordless cycle.
       \end{lemma}
       
       \begin{lemma}
              With the set of chordless paths, one can find all paths in the graph.
       \end{lemma}

       The detailed proofs and additional visualisations are provided in the Appendix \ref{p:3.4}.

       \begin{theorem}
Chordless cycles and chordless paths together provide enough information to describe the structure of a graph.
\end{theorem}

The result follows from the preceding lemmas. Thus, the decomposition of the graph is now complete.

    \section{Chordless Structure based GNN: architecture and expressiveness}
       In this section, we introduce the architecture of CSGNN, and show the expressiveness
       of CSGNN by giving provable evidence.
       \subsection{Model Architecture}
       The architecture of CSGNN is similar to Message Passing Neural Networks (MPNNs),
       but with a new aggregation method and spread edges. The overall
       architecture of CSGNN is as follows:
       \begin{align*}
h_{CP}^{t}(v) = \mathrm{AGG_1}\left( \{ h_{u}^{(t)} : u \in CP_{k}(v) \} \right) & \quad
h_{CC}^{t}(v) = \mathrm{AGG_2}\left( \{ h_{u}^{(t)} : u \in CC_{k}(v) \} \right) \\
h_{v}^{(t+1)} &= h_{CP}^{t}(v) \oplus h_{CC}^{t}(v)
\end{align*}

       The implementation above is inherited from the message-passing template.
       For a more visible representation, we can consider another implementation
       by encoding the chordless cycles into the representations of the nodes.
       \begin{align*}
h_{v}^{(0)} = x_{v} \oplus \mathrm{ENCODE}(CC_{k}(v)) \quad
h_{v}^{(t+1)} = \mathrm{AGG}\left( \{ h_{u}^{(t)} : u \in CP_{k}(v) \} \right)
\end{align*}

       \subsection{Theoretical Background}\label{sec:4.2}
       For further analysis, we will introduce some definitions in the section to
       quantify the expressiveness of GNNs.
       \begin{definition}
              A GNN $GNN_1$ is more expressive (not strictly) than another GNN $GNN_2$ if $GNN_1$ can distinguish every pair of the graphs that $GNN_2$ can distinguish. We denote this relation as $GNN_2 \preccurlyeq GNN_1$.
       \end{definition}
       In order to make analysis easier, we can consider a new definition with intuition from coloring problems.
       By considering a subset of the graphs with good properties instead of of the graphs, we can head to our result more efficiently.
       \begin{definition}
              A graph is said to be $1-WL-colorable$ if there is no two vertices
              sharing the same edge have the same color after the $1-WL$ coloring.\par
       \end{definition}
       It is clear that not each graph is $1-WL-colorable$, but we can still transform the graphs in $1-WL-colorable$ format, for proving our theory's robustness.
       We can consider the following lemma, to uniform the form of the graphs' format, to get conclusions easier.
       \begin{lemma}\label{vis}
              For a finite set of graphs, each of them can be map in the form of $1-WL-colorable$ graph.\par
       \end{lemma}
       The detailed proof and visual interpretation are provided in Appendix~\ref{p:4.3}.

       \par
       
       \begin{definition}
              For two graphs $G_{1}$ and $G_{2}$, if they have the same colored
              form(under isomorphism) after 1-WL coloring, we say they are
              1-WL-equivalent, and denote $G_1 \sim_{1-WL}G_2$.\par
       \end{definition}\par
       \begin{lemma}
              \label{minimal} For two graphs $G_{1}$ and $G_{2}$, if they are
              $1-WL-equivalent$ but not isomorphic, there exist minimal subgraphs
              $G_{1}'$ and $G_{2}'$ of $G_{1}$ and $G_{2}$ such that $G_{1}'$ and
              $G_{2}'$ are not isomorphic but $G_1^{'} \sim_{1-WL}G_2^{'}$.
       \end{lemma}\par
        The proof of the lemma is obvious since the minimal subgraphs can be itself
              whatever.
       Hence, in order to generate the structure of the graph after 1-WL coloring,
       we can start with the nodes in the same color and then add colors
       adjacent to the previous layer, layer by layer based on the color. But
       the structure of the graph is complex, we turn to consider the minimal
       subgraph that is defined in the Lemma \ref{minimal}, and consider a more
       common form in which each layer has only one color, estimate whether each
       graph is $1-WL-colorable$ in the follow-up analysis.\par
       \begin{figure}[h]
    \centering

    \begin{subfigure}[b]{0.48\textwidth}
        \centering
        \includegraphics[width=1.2\textwidth, trim=4cm 6cm 0cm 6cm, clip]{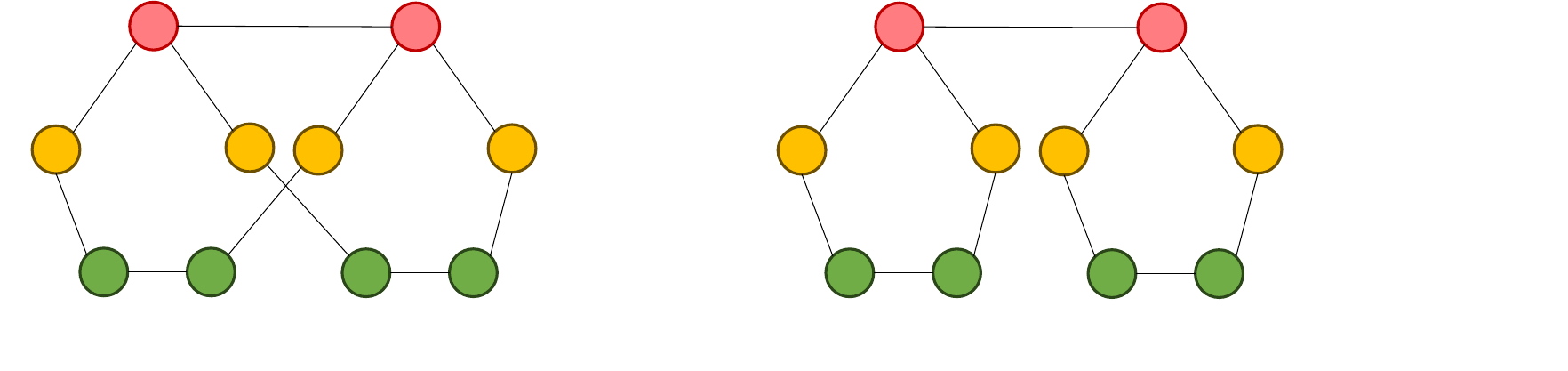}
        \caption{
        Two WL-equivalent graphs can be distinguished through chordless cycles.}
        \label{fig:sub1}
    \end{subfigure}
    \hfill
    \begin{subfigure}[b]{0.48\textwidth}
        \centering
        \includegraphics[width=0.9\textwidth, trim=0cm 6cm 0cm 4cm, clip]{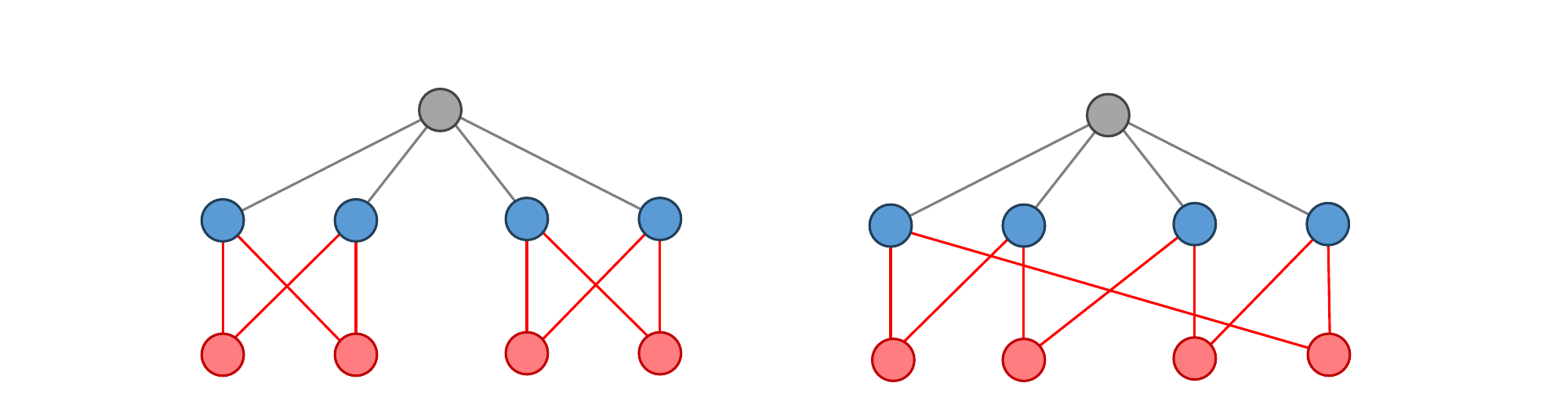}
        \includegraphics[width=0.9\textwidth, trim=0cm 6cm 0cm 6cm, clip]{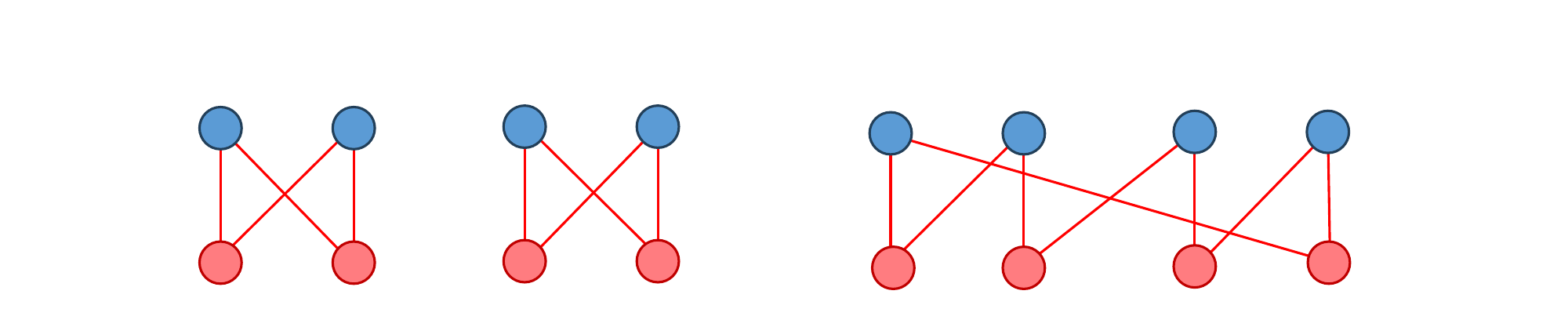}
        \caption{
        Two pairs of WL-equivalent graphs that are distinguishable by CSGNN.}
        \label{bi}
    \end{subfigure}

    \caption{
    Examples of WL-indistinguishable graphs where CSGNN succeeds: (a) distinction by chordless cycles; (b) distinction via component structures.}
    \label{fig:csgnn_examples}
\end{figure}

       In order to describe CSGNN's expressiveness, we need to introduce the concept below, which is similar to the definition of graph isomorphism naturally.\par
       \begin{definition}
              A colored graph $G=(V,E,c)$ is $color-equivalent$ if for any pair of
              nodes $u_{1}$,$u_{2}$ of the same color, there exists a map
              $\phi :V\rightarrow V$ that satisfies the following conditions:
              \begin{itemize}[itemsep=2pt, parsep=0pt, topsep=0pt]
                     \item $\phi$ is a bijection which satisfies $\phi(u_{1})=u_{2}$,
                            and $\phi(u_{2})=u_{1}$.

                     \item For any $v_{1},v_{2}\in V$, $v_{1}$ and $v_{2}$ are
                            adjacent if and only if $\phi(v_{1})$ and
                            $\phi(v_{2})$ are adjacent.

                     \item For any $v\in V$, the color of $v$ is the same as the
                            color of $\phi(v)$.
              \end{itemize}
       \end{definition}

       \subsection{Expressiveness of CSGNN}
       First, by showing the expressiveness based on aggregating the chordless
       paths, we can get a lower bound of the expressiveness of CSGNN.

       \begin{theorem}
              CSGNN is more expressive than the k-hop GNN (KPGNN) with $CP_{[k]}$.
       \end{theorem}\par
       \begin{proof}
              Since every chordless path of length $\leq k$ is within the computational scope of $KPGNN$, we have 
              $CP_{[k]}\preccurlyeq KPGNN$, we only need to prove the other direction. By considering theorem \ref{thm1}, we
              can get the proof of the theorem.
       \end{proof}
       \par Furthermore, to observe the benefit of the chordless
       cycle, we can consider the following lemma.

       \begin{lemma}
              If $G_{1},G_{2}$ are two non-isomorphic bipartite graphs that have different amounts of connected components, 
              then CSGNN can distinguish $G_{1}$ and $G_{2}$. \label{bilemma}
       \end{lemma}
       In some special cases, we can consider the following example \ref{bi}, showing
       the hybrid capability of considering minimal graphs in Lemma \ref{minimal} and
       using the Lemma \ref{bilemma}. \par
       
       For a more general case, we limit the condition of the graphs. Because we
       have realized that the upper bound of standard GNNs is caused by cycles. We
       try to construct the graph first by formatting from a node to a tree by layers,
       and finally, we form the cycle in the last layer. The process aims to constrain
       regular representations of graphs. In other words, the reasonableness can
       be shown, since two non-isomorphic graphs must have different nodes, and we
       form the graph from the different nodes, which also matches the minimal
       subgraph in Lemma 3.6.

       \begin{theorem}
              Let $G_{1},G_{2}$ be two non-isomorphic graphs satisfies the following conditions:
              \begin{itemize}[itemsep=2pt, parsep=0pt, topsep=0pt]
                     \item $G_{1}$ and $G_{2}$ are $color-equivalent$

                     \item $G_{1}$ and $G_{2}$ have different amounts of connected components.
              \end{itemize}
              Then CSGNN can distinguish $G_{1}$ and $G_{2}$.
       \end{theorem}
       The proof is left at appendix \ref{p:4.9}.

       Notice that the cut-edge is a symbol of components, we have the following
       corollary.

       \begin{corollary}
              CSGNN is cut-edge sensitive.
       \end{corollary}
       From the content above, we get the expressiveness of CSGNN. GNN has
       limited expressiveness since graphs don't have a uniform structure. But after
       processing by CSGNN, we can get a more regular form of the graph, which improves
       the expressiveness of GNNs, see Figure \ref{uniform} for an example.

       \subsection{Feature Encoding}\label{4.4}
       To realize the topological structure of the graph better, we can introduce
       the concept of \textbf{regular cell complex} from \cite{hansen2019toward}, and get inspiration from design
       \begin{definition}
              A \textbf{regular cell complex} is a topological space $X$ with a
              partition into subspaces $\{X_{\alpha}\}_{\alpha \in P_X}$
              satisfying the following conditions:
              \small
              \begin{itemize}[itemsep=2pt, parsep=0pt, topsep=0pt]
                     \item $\forall x \in X$, every sufficiently small
                            the neighborhood of $x$ intersects only finitely many
                            $X_{\alpha}$.

                     \item $\forall \alpha,\beta \quad \bar{X_\alpha}\cap X_{\beta}
                            \neq \emptyset$ only if $X_{\beta}\subseteq \bar{X_\alpha}$

                     \item Every $X_{\alpha}$ is homeomorphic to $\mathbf{R}^{n_\alpha}$
                            for some $n_{\alpha}$.

                     \item For every $\alpha$, there is a homeomorphism of a
                            closed ball in $\mathbf{R}^{n_\alpha}$ to
                            $\bar{X_\alpha}$ that maps the interior of the ball
                            homomorphically onto $X_{\alpha}$
              \end{itemize}
       \end{definition}
       
       Here we can consider $P_{X}$ denotes the set of the minimal cycles in the graph, since the intersections
       of the minimal cycles are always empty set.\par
\begin{figure}[t]
    \centering

    \begin{subfigure}[t]{0.5\textwidth}
        \centering
        \includegraphics[width=0.85\textwidth]{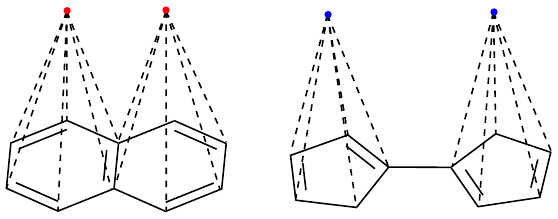}
        \caption{
        Topological cones on Decalin and Bicyclopenty, two 1-WL indistinguishable molecules.}
        \label{fig:cone}
    \end{subfigure}
    \hfill
    \begin{subfigure}[t]{0.45\textwidth}
        \centering
        \includegraphics[width=0.85\textwidth]{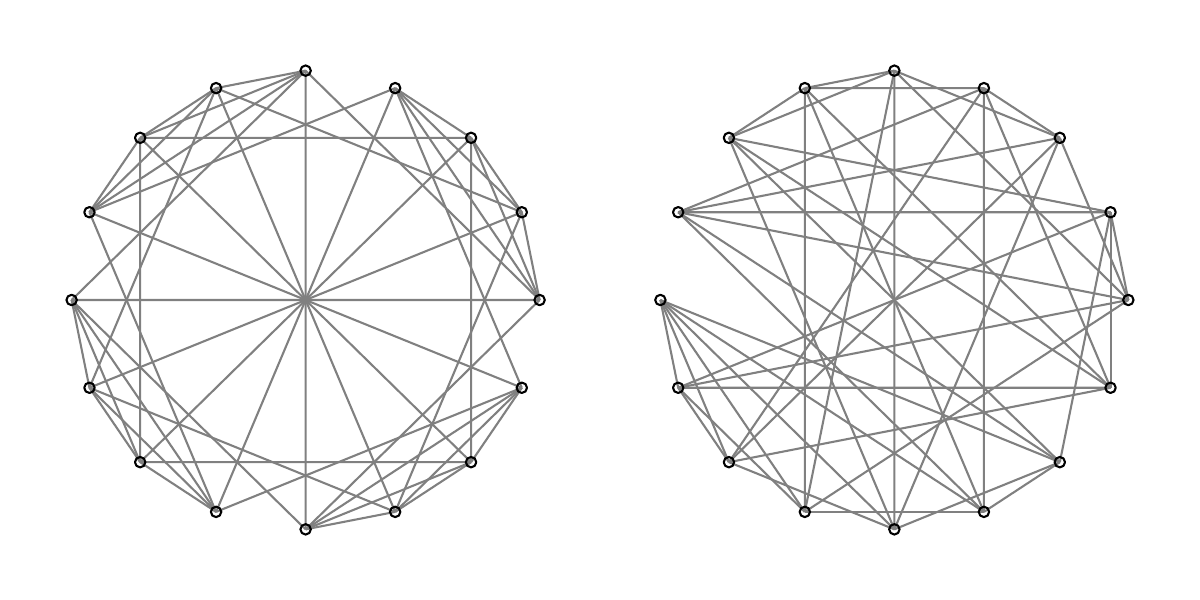}
        \caption{
        Two 3-WL indistinguishable graphs from SR(16,6,2,2).}
        \label{fig:sr22}
    \end{subfigure}

    \caption{
    Examples of non-isomorphic graphs indistinguishable by WL tests. (a) 1-WL failure on molecular graphs; (b) 3-WL failure on strongly regular graphs.
    \vspace{-.3in}
    }
    \label{fig:wl_failures}
\end{figure}

 We can encode the structure
       with an intuition from chemistry. In some ways, the nodes on the cycle compose
       an equivalent class (example: carbon on the benzene ring have similar properties),
        so we want the encoding to be a mapping from the
       original topological space to a quotient space, which corresponds to the concept
       of \textbf{Topological cone}
       \begin{definition}
              For a topological space $X$, the \textbf{topological cone} of $X$ is
              the quotient space
              $CX:=(X\times I)/(X\times \{1\})$, where $I$ denotes the interval $[0,1]$.
       \end{definition}
       So the encoding can be realized as a mapping from the subspaces of the
       regular cell to the topological cone of the graph such as in Figure
       \ref{fig:cone}. Hence the feature can be encoded as the following:
       \[
              Feature_{i}(u)=\mathbf{1}\Big[\exists CC_{i}\in X,\quad s.t.\quad u
              \in CC_{i}\Big]
       \]where $u$ is the node in the graph, and $CC_{i}$ is the chordless cycle(always
       minimal cycle) in the graph, and $\mathbf{1}$ is the indicator function. So
       with the encoding, we can get an equivalent class on the cycle since all
       the node on the cycle is encoded in 1. The model can be realized as a
       mapping $M \rightarrow CM$ with more information from the topological
       structure of the graph, where $M$ is the original structure. While the structural features in \cite{sgcgnn} rely on symmetry group counts, our approach emphasizes preserving the structural similarity of nodes within cyclic (ring) substructures.
       \par In another way, we can consider the encoding as
       a binary molecular Fingerprint\cite{mcgregor1997clustering} embedding
       into the node features. And our model performs well in the
chemical graph data sets in the experiments, which are shown in the following section.

\section{Experiments}

\begin{table}[h]  %
  \vspace{-2pt}                     %
  \centering
  \small
  \begin{tabular}{|c|c|c|c|}
    \hline
    \textbf{Dataset} & \textbf{Task}     & \textbf{Size/Type}             & \textbf{Special Property}             \\
    \hline
    GRAPH8C          & Isomorphism       & 11117 graphs (8 nodes)         & All non-isomorphic graphs             \\
    \hline
    EXP              & Isomorphism       & 600 pairs                      & 1-WL indist. / 2-WL dist.             \\
    \hline
    SR               & Isomorphism       & 105 pairs (25 nodes)           & 3-WL indist.                          \\
    \hline
    ZINC             & Regression        & 250K molecules                 & Predict chemical properties           \\
    \hline
    TUDataset        & Classification    & 120+ datasets                  & Chem. / social graphs                 \\
    \hline
    ogbg-MOLHIV      & Classification    & 40K+ molecules                 & MoleculeNet; HIV activity             \\
    \hline
  \end{tabular}
  \caption{Overview of datasets.}
  \label{tab:dataset1}
  \vspace{-10pt}                     %
\end{table}

       To qualify the expressiveness of CSGNN in distinguishing graphs, we apply CSGNN on the synthetic dataset in Section~\ref{5.1} and real-world datasets in Section~\ref{5.2}, which include graph isomorphism tasks, regression tasks, and classification tasks. The detailed experimental setup is provided in Appendix~\ref{setup}, and the detailed dataset introduction is in Appendix \ref{dataset}.

       \subsection{Synthetic Datasets}\label{5.1}
         \textbf{Results}
       The performance is given by the table and the figure, we find that CSGNN performs
       well on graph isomorphism tasks. According to the results of SR datasets,
       CSGNN has stronger expressiveness than 3-WL. Additionally, by running the
       code in \cite{balcilar2021breaking} and checking their results, we can
       find that the expressiveness of fundamental GNNs is limited by 1-WL test,
       then less than 3-WL test. Above all, our experiments show that CSGNN has excellent
       expressiveness.

\begin{table}[H]
\centering
\begin{subtable}[t]{0.48\textwidth}
\vspace{0pt}
    \centering
    \small
    \begin{tabular}{|lll|}
        \hline
        \textbf{Model}                            & \textbf{EXP $\downarrow$} & \textbf{Graph8c $\downarrow$} \\
        \hline
        GCN \cite{kipf2017} &600& $7.727*10^{-5}$\\[0.5pt]
        GAT \cite{velivckovic2017graph}           & 600                       & $2.958*10^{-5}$               \\[0.5pt]
        GIN\cite{limitation}                      & 600                       & $6.247*10^{-6}$               \\[0.5pt]
        GNNML1\cite{balcilar2021breaking} & 600 & $5.389*10^{-6}$\\[0.5pt]
        PathNN\cite{pathnn}                       & 0                         & N/A                           \\[0.5pt]
        ChebNet\cite{defferrard2016convolutional} & 0                         & $7.120*10^{-7}$               \\[0.5pt]
        PAIN\cite{pathGNNs2024} & 0 &N/A\\
        \hline
        CSGNN                                     & 0                         & 0                             \\
        \hline
    \end{tabular}
    \caption{Undistinguished pairs in EXP and failure rate on Graph8c.} 
    \label{tab:exp_graph8c}
\end{subtable}
\hfill
\begin{subtable}[t]{0.48\textwidth}
\vspace{0pt}
    \centering
    \small
    \begin{tabular}{|clcc|}
        \hline \textbf{No.} & \textbf{Model}                            & \textbf{Test MAE$\downarrow$}   & \textbf{T/A} \\
        \hline
        1            & ESA\cite{2402.10793}            & $0.0122^{(0.0004)}$ & \checkmark                           \\
        2                     & TIGT\cite{choi2024topology}     & $0.014$             & \checkmark                           \\
        \hline
        3                     & CSGIN (Ours)                     & $0.0216$            &                                      \\
        \hline
        4                     & GRIT\cite{ma2023graph}          & $0.023$             & \checkmark                           \\
        5                     & GraphGPS\cite{rampavsek2022recipe} & $0.024^{(0.007)}$   & \checkmark                       \\
        6                     & SignNet\cite{lim2022sign}       & $0.024^{(0.003)}$   & \checkmark                           \\
        7                     & Graphormer\cite{ying2021transformers} & $0.036^{(0.002)}$ & \checkmark                   \\
        8                     & $\delta$-2-GNN\cite{morris2020weisfeiler} & $0.042^{(0.003)}$ &                             \\
        \hline
    \end{tabular}
    \caption{Test MAE on ZINC dataset. T/A = Transformer/Attention-Based.}
    \label{tab:zinc_mae}
\end{subtable}
\caption{
Comparative evaluation of GNN expressiveness and real-world performance.
(a) Number of indistinguishable pairs on the EXP dataset and failure rate on Graph8c (lower is better).
(b) Test MAE on the ZINC dataset, sorted by performance. Models marked with \checkmark are Transformer/Attention-based.
}

\end{table}

       \begin{figure}[h]
  \centering
\includegraphics[width=0.8\textwidth, trim=0 0 0 30, clip]{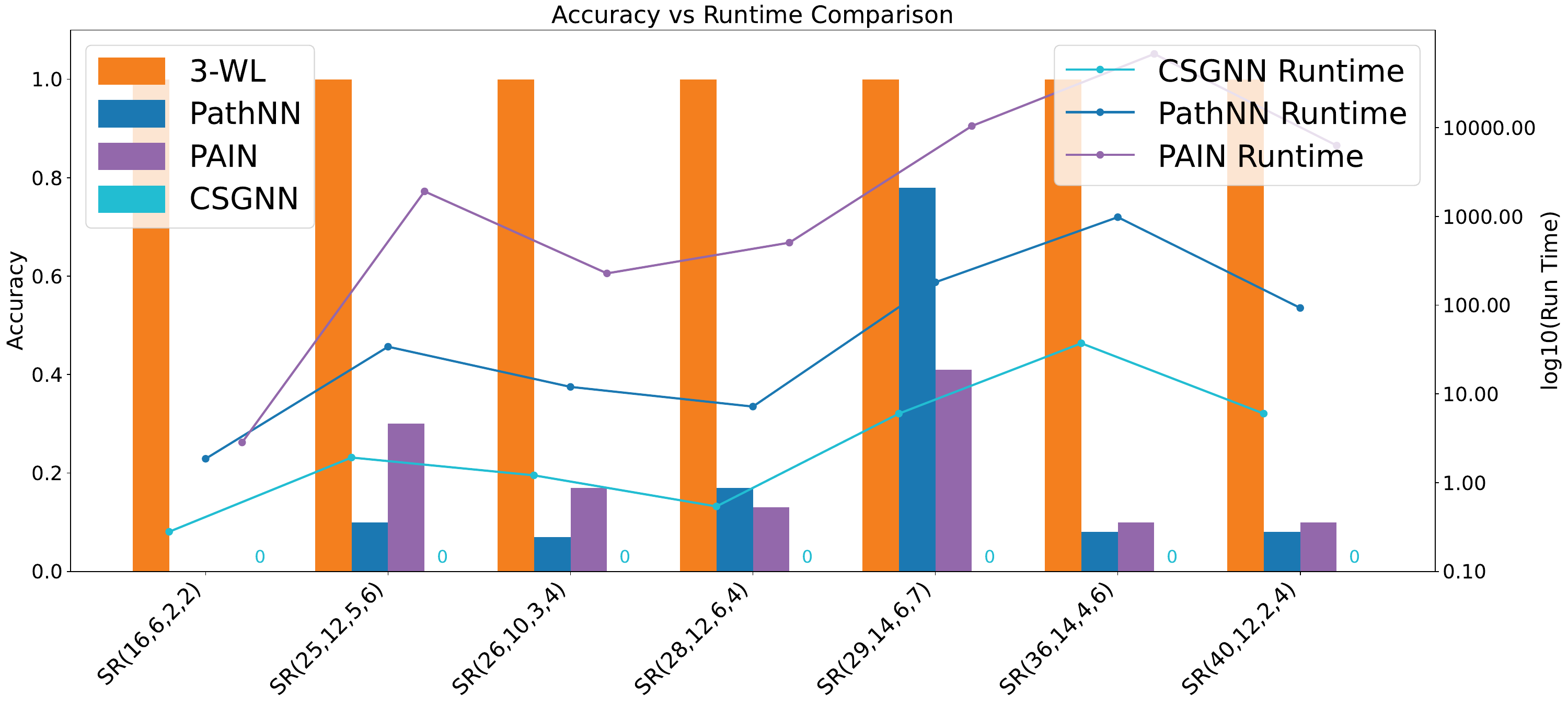}
\caption{Accuracy and Run time(s) comparison on SR datasets}
\end{figure}
\vspace{-25pt}
\subsection{Real-World Datasets}\label{5.2}       
\textbf{Result} According to the benchmark table from \url{https://paperswithcode.com/sota/}, our CSGNNs achieve strong performance\footnote{We merged the results of ESA and retained the best one} on the full ZINC dataset \ref{tab:zinc_mae}. Compared to CY2C-GCN \cite{choi2022cycle}, our models are more robust and resource-efficient, notably avoiding out-of-memory errors on the DD dataset when tested on an A100 GPU (see Table~\ref{tab:sub1}). Moreover, compared to PathNN, our model significantly reduces memory usage, as shown in Table 4. Our method outperforms many classical models on the ogbg-MolHIV dataset, as shown in Table \ref{tab:sub2}.

\begin{table}[H]
  \centering
  \small
  \begin{subtable}[t]{0.60\textwidth}
    \vspace{0pt}  %
    \centering
    \resizebox{\textwidth}{!}{
      \begin{tabular}{lcccc}
        \toprule
                     & DD                      & PROTEINS              & IMDB-B                & IMDB-M                \\
        \midrule
        DGCNN\cite{zhang2018end}   & 76.6 $\pm$ 4.5          & 72.9 $\pm$ 3.5        & 69.2 $\pm$ 3.0        & 45.6 $\pm$ 3.4        \\
        DiffPool\cite{ying2018hierarchical} & 75.0 $\pm$ 4.3 & 73.7 $\pm$ 3.5        & 68.4 $\pm$ 3.3        & 45.6 $\pm$ 3.4        \\
        ECC \cite{simonovsky2017dynamic}    & 72.6 $\pm$ 4.1          & 72.3 $\pm$ 3.4        & 67.7 $\pm$ 2.8        & 43.5 $\pm$ 3.1        \\
        GIN \cite{limitation}               & 75.3 $\pm$ 2.9          & 73.3 $\pm$ 4.0        & 71.2 $\pm$ 3.9        & 48.5 $\pm$ 3.3        \\
        GraphSAGE \cite{hamilton2018}       & 72.9 $\pm$ 2.0          & 73.0 $\pm$ 4.5        & 68.8 $\pm$ 4.5        & 47.6 $\pm$ 3.5        \\
        GAT \cite{velivckovic2017graph}     & 73.9 $\pm$ 3.4          & 70.9 $\pm$ 2.7        & 69.2 $\pm$ 4.8        & 48.2 $\pm$ 4.9        \\
        SPN \cite{abboud2022shortest}       & 77.4 $\pm$ 3.8          & 74.2 $\pm$ 2.7        & N/A                   & N/A                   \\
        PathNet \cite{sun2022beyond}        & OOM                     & 70.5 $\pm$ 3.9        & 70.4 $\pm$ 3.8        & 49.1 $\pm$ 3.6        \\
        Nested GNN \cite{zhang2021nested}   & 77.8 $\pm$ 3.9          & 74.2 $\pm$ 3.7        & N/A                   & N/A                   \\
        PathNN \cite{pathnn}                & 77.0 $\pm$ 3.1          & 75.2 $\pm$ 3.9        & 72.6 $\pm$ 3.3        & \textbf{50.8 $\pm$ 4.5} \\
        CY2C-GCN \cite{choi2022cycle}       & OOM                     & 61.0 $\pm$ 5.3        & 58.0 $\pm$ 4.1        & 35.3 $\pm$ 2.1        \\
        KAGNN \cite{bresson2024kagnn}       & 73.3 $\pm$ 4.5          & \textbf{75.6 $\pm$ 3.6} & 72.6 $\pm$ 4.5        & 50.0 $\pm$ 4.1          \\
        \midrule
        CSGNN                               & \textbf{78.1 $\pm$ 2.6} & 75.4 $\pm$ 3.4        & \textbf{72.8 $\pm$ 3.6} & 50.6 $\pm$ 4.1          \\
        \bottomrule
      \end{tabular}
    }
    \caption{Performance comparison across DD, PROTEINS, IMDB-B, and IMDB-M datasets. OOM = out-of-memory; N/A = not available.}
    \label{tab:sub1}
  \end{subtable}
  \hfill
  \begin{subtable}[t]{0.35\textwidth}
    \vspace{0pt}  %
    \centering
    \small
    \begin{tabular}{lc}
        
      \toprule
      \textbf{Model} & \textbf{ROC-AUC} \\
      \midrule
      GIN         & 75.58 $\pm$ 1.40 \\[1pt]
      GCN         & 76.06 $\pm$ 0.97 \\[1pt]
      ESAN        & 78.00 $\pm$ 1.22 \\[1pt]
      CIN         & 80.94 $\pm$ 0.57 \\[1pt]
      GSN         & 77.99 $\pm$ 1.00 \\[1pt]
      Graphormer  & 80.51 $\pm$ 1.06 \\[1pt]
      GPS         & 78.80\\
      pathNN     & 79.17 $\pm$ 1.09 \\[1pt]
      PAIN        & 78.50 $\pm$ 1.40 \\[2pt]
      \hline
      CSGIN       & 82.22          \\
      \bottomrule
    \end{tabular}
    \caption{ROC-AUC results on the ogbg-MolHIV dataset.}
    \label{tab:sub2}
  \end{subtable}

  \caption{(a) Performance comparison on four benchmark datasets; (b) ROC-AUC results for ogbg-MolHIV.}
  \label{tab:combined}
  \end{table}

    \section{Conclusion}
       In this paper, we observe that chord is a key factor in achieving simple and
       expressive GNNs. Accordingly, we propose a new GNN framework, CSGNN, which
       is based on chordless structures. Theoretical analysis shows that CSGNN
       achieves expressiveness beyond KPGNN but only with
       $\mathcal{O}(c(G)(|V|+| E|))$ complexity. Extensive experiments demonstrate
       that CSGNN outperforms existing GNN models by embedding cycles' feature
       into graph but only introduces $\mathcal{O}(|CC|(|V|+|E|))$ additional
       computational cost.
\\

    \bibliographystyle{plain}
    \bibliography{nips/example_paper.bib}
    \appendix
    \newpage
\section*{NeurIPS Paper Checklist}

\begin{enumerate}

\item {\bf Claims}
    \item[] Question: Do the main claims made in the abstract and introduction accurately reflect the paper's contributions and scope?
    \item[] Answer: \answerYes{} %
    \item[] Justification: The claims in the abstract and introduction accurately reflect the scope and contributions of the paper, as detailed in Sections 1.

\item {\bf Limitations}
    \item[] Question: Does the paper discuss the limitations of the work performed by the authors?
    \item[] Answer: \answerNo{} %
    \item[] Justification: The limitations of our work are discussed at the points in the paper where they naturally arise.

\item {\bf Theory assumptions and proofs}
    \item[] Question: For each theoretical result, does the paper provide the full set of assumptions and a complete (and correct) proof?
    \item[] Answer: \answerYes{} %
    \item[] Justification: All theoretical results are stated with their full assumptions and accompanied by complete proofs in section 4 and the appendix.

    \item {\bf Experimental result reproducibility}
    \item[] Question: Does the paper fully disclose all the information needed to reproduce the main experimental results of the paper to the extent that it affects the main claims and/or conclusions of the paper (regardless of whether the code and data are provided or not)?
    \item[] Answer: \answerYes{} %
    \item[] Justification: We provide full implementation details in Appendix to enable reproduction of all key experiments.

\item {\bf Open access to data and code}
    \item[] Question: Does the paper provide open access to the data and code, with sufficient instructions to faithfully reproduce the main experimental results, as described in supplemental material?
    \item[] Answer: \answerYes{} %
    \item[] Justification:  We include anonymized link to our code and data in the footnote, along with detailed instructions to reproduce results.

\item {\bf Experimental setting/details}
    \item[] Question: Does the paper specify all the training and test details (e.g., data splits, hyperparameters, how they were chosen, type of optimizer, etc.) necessary to understand the results?
    \item[] Answer: \answerYes{} %
    \item[] Justification: All relevant training and evaluation details, including hyperparameters and data splits, are described in Section 5 and Appendix.

\item {\bf Experiment statistical significance}
    \item[] Question: Does the paper report error bars suitably and correctly defined or other appropriate information about the statistical significance of the experiments?
    \item[] Answer: \answerYes{} %
    \item[] Justification: We report standard deviation over runs for each metric and explain the computation method in Section 5.

\item {\bf Experiments compute resources}
    \item[] Question: For each experiment, does the paper provide sufficient information on the computer resources (type of compute workers, memory, time of execution) needed to reproduce the experiments?
    \item[] Answer: \answerYes{} %
    \item[] Justification: We specify the hardware and compute time used for each experiment in Section 5 and Appendix.
    
\item {\bf Code of ethics}
    \item[] Question: Does the research conducted in the paper conform, in every respect, with the NeurIPS Code of Ethics \url{https://neurips.cc/public/EthicsGuidelines}?
    \item[] Answer: \answerYes{} %
    \item[] Justification: Our work adheres fully to the NeurIPS Code of Ethics, and no ethical concerns are raised by our methodology or data usage.

\item {\bf Broader impacts}
    \item[] Question: Does the paper discuss both potential positive societal impacts and negative societal impacts of the work performed?
    \item[] Answer: \answerNA{} %
    \item[] Justification: There is no societal impact of the work performed.
    
\item {\bf Safeguards}
    \item[] Question: Does the paper describe safeguards that have been put in place for responsible release of data or models that have a high risk for misuse (e.g., pretrained language models, image generators, or scraped datasets)?
    \item[] Answer: \answerNA{} %
    \item[] Justification: The paper poses no such risks.

\item {\bf Licenses for existing assets}
    \item[] Question: Are the creators or original owners of assets (e.g., code, data, models), used in the paper, properly credited and are the license and terms of use explicitly mentioned and properly respected?
    \item[] Answer: \answerYes{} %
    \item[] Justification: All datasets and code used are properly cited with licensing terms.

\item {\bf New assets}
    \item[] Question: Are new assets introduced in the paper well documented and is the documentation provided alongside the assets?
    \item[] Answer: \answerYes{} %
    \item[] Justification: We release a new code along with complete documentation, licensing, and usage instructions in the repository provided in the footnote.

\item {\bf Crowdsourcing and research with human subjects}
    \item[] Question: For crowdsourcing experiments and research with human subjects, does the paper include the full text of instructions given to participants and screenshots, if applicable, as well as details about compensation (if any)? 
    \item[] Answer: \answerNA{} %
    \item[] Justification:  The paper does not involve crowdsourcing nor research with human subjects.

\item {\bf Institutional review board (IRB) approvals or equivalent for research with human subjects}
    \item[] Question: Does the paper describe potential risks incurred by study participants, whether such risks were disclosed to the subjects, and whether Institutional Review Board (IRB) approvals (or an equivalent approval/review based on the requirements of your country or institution) were obtained?
    \item[] Answer: \answerNA{} %
    \item[] Justification: The paper does not involve crowdsourcing nor research with human subjects.

\item {\bf Declaration of LLM usage}
    \item[] Question: Does the paper describe the usage of LLMs if it is an important, original, or non-standard component of the core methods in this research? Note that if the LLM is used only for writing, editing, or formatting purposes and does not impact the core methodology, scientific rigorousness, or originality of the research, declaration is not required.
    \item[] Answer: \answerNA{} %
    \item[] Justification: The core method development in this research does not involve LLMs as any important, original, or non-standard components.

\end{enumerate}

    \section{Related Work}
       \textbf{GNNs that use paths} As \cite{pathnn} wrote in their paper, Graphormer\cite{ying2021transformers},
       PEGN\cite{zhao2020persistence}, SP-MPNN\cite{abboud2022shortest}, Geodesic
       GNN\cite{kong2022geodesic} use the shortest paths to embed the graph
       structure, which aims to use fewer paths to represent the graph structure.
       There are some other GNNs that use specific paths to represent the
       graph structure. \cite{pathnn} introduce PathNN that uses the sp-path, \cite{pathGNNs2024} develop PathNN  focus on single shortest paths, all shortest paths and all simple paths, and \cite{liu2019geniepath} uses GeniePath adaptively as receptive paths for
       breadth/depth exploration and filter useful/noisy signals. \cite{truong2024weisfeiler}
       introduces topological message-passing scheme PCN(Path Complex Networks)
       operating on path complexes and a novel graph isomorphism test PWL, with the
       the theoretical connection between PWL and the latter higher-order WL tests.

       \textbf{GNNs that use cycles} At the same time, starting from the subgraph count
       GNN, cycle as an important subgraph, has been widely used in graph representation.
       \cite{pmlr-v162-yan22a} uses the SPT cycle bases family to learn cycle
       representations to enhance the expressiveness of GNN for inductive
       relation prediction tasks in knowledge graphs.\cite{cwn} proposes a new
       message-passing aggregation method based on cell complexes(including
       cycles) to enhance the expressiveness of GNNs. \cite{choi2022cycle}
       introduces a way of using cycle structures in aggregation and shows its
       expressiveness by universal covers. They connect the nodes in the same cycle
       and then aggregate them as cliques in some additional GNN layers. \cite{yan2024cycle}
       proposes a novel edge structure encoding CycleNet based on the position of
       the edge in the cycle spaces.

       \textbf{The limitation(expressiveness) of MPNNs} It has shown that
       \cite{limitation} MPNNs are at most as powerful as the WL test in
       distinguishing isomorphic graphs. On the other hand, the limitation leads
       to a lack in counting simple substructures\cite{chen2020can} like cycles.
       At the same time, there is a lot of work that tries to enhance the expressiveness
       of GNNs by using graph features, graph topology, and enhancing the architecture
       of GNNs, see the survey\cite{zhang2023expressive} for more details.

\section{Experimental Setup}
\label{setup}
\subsection{Experimental Setup on synthetic datasets}
\textbf{Experimental
       setup} Following \cite{pathnn}, for all experiments in synthetic datasets,
       the aggregation function is set to the identity function. We set initial
       node features to be vectors of ones and process them with MLPs.\par For the
       SR, EXP, and graph8c datasets, we investigate whether the proposed models
       have the right inductive bias to distinguish the graphs. Like \cite{errica2019fair}
       , we consider two graphs to be non-isomorphic if the Euclidean distance between
       their representations is not below a threshold $\epsilon$, where the
       presentation is generated by an untrained model. And we set the radius to $3$ for
       detecting cycles in use. Then we use some cycle-convolution layers to
       aggregate and apply a GCN layer and a linear layer at last. \par
\subsection{Experimental Setup on real-world datasets}
\textbf{Experimental Setup} For regression tasks on ZINC, we add the feature introduced in Section~\ref{4.4} to the node features, followed by a linear layer for encoding. We apply simple MLP models with various convolution layers, using batch normalization between them and an LSTM before the final linear layer.

For classification tasks on TUDatasets, we incorporate cycle convolutions similar to \cite{choi2022cycle}, aggregating nodes along chordless cycles step by step based on their distance in the original graph. This helps the model differentiate structures like ortho, meta, and para. We also use GAT layers based on MPNNs, where edge indices are given by chordless paths.

For classification on ogbg-MOLHIV, we add the feature from Section~\ref{4.4} into a GIN model. We follow the evaluation protocol of \cite{errica2019fair} on TUDatasets, using 10-fold cross-validation and predefined data splits. Hidden dimensions are chosen from $\{32, 64\}$, and dropout rates from $\{0, 0.5\}$ are applied to the final MLP. For all datasets, we use a consistent batch size of 1. We also test various convolution strategies, including path-first and cycle-first approaches, to systematically assess their impact.
\section{Statistics of Chordless Cycles in various datasets}

       \begin{figure}[H]
              \centering
              \includegraphics[width=0.5\textwidth, trim=0cm 4cm 0cm 4cm, clip]{
                     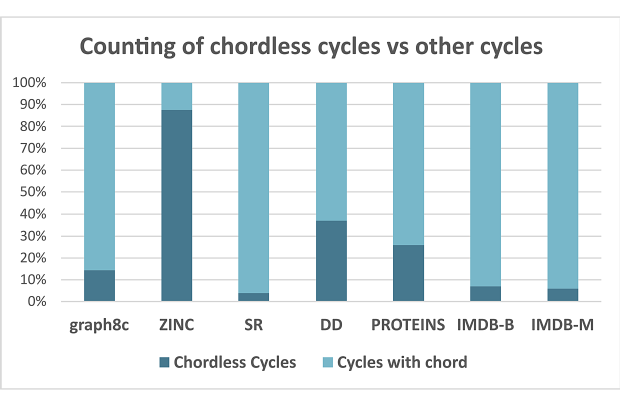
              }
              \caption{By comparing the counts between chordless cycles and
              other cycles, we can see that the chordless cycles demonstrate superior
              performance, achieving significantly lower costs }
              \label{nature}
       \end{figure}
    \begin{table}[H]
    \centering
    \begin{tabular}{lrr}
      \toprule
      Dataset   & CSGNN & PathNN \\
      \midrule
      Protein   & 248   & 408    \\
      DD        & 704   & 2664   \\
      IMDB-B    & 386   & 390    \\
      IMDB-M    & 584   & 444    \\
      \bottomrule
      
    \end{tabular}
    \label{mem}
    \captionof{table}{Model Memory Usage (MB)}
    \end{table}

\section{Dataset introduction}\label{dataset}
\subsection{Synthetic Datasets}
\textbf{Dataset Overview} We evaluate the performance of our CSGNN in three
       publicly available datasets. \textbf{Graph8c} consists of all the
       11117 possible connected non-isomorphic simple graphs with 8 nodes.
       \textbf{EXP dataset} \cite{abboud2020surprising}includes 600 pairs of 1-WL
       equivalent graphs which is distinguishable under the 2-WL algorithm.
       \textbf{SR dataset} \footnote{EXP and SR dataset can be found in
       \url{https://users.cecs.anu.edu.au/~bdm/data/graphs.html}}consists of
       strongly regular graphs which is undistinguishable under the 3-WL
       algorithm. Where each graph in SR25 has 25 nodes, and contains 15
       graphs generating 105 pairs for isomorphism comparison.\par
\subsection{Real-World Datasets}
\textbf{Dataset Overview} We evaluate CSGNN in real-world tasks, including
       regression tasks and classification tasks in three datasets.\par \textbf{ZINC}\cite{irwin2012zinc}
       It is a molecular graph regression dataset with 250,000 molecules, used
       for predicting molecular properties, and ZINC(subset) takes a 12k subset
       of it.\par \textbf{TUDatasets}\cite{morris2020tudataset} It is a
       collection of over 120 datasets for graph classification, covering various
       applications such as chemistry and bioinformatics. \par
       \textbf{ogbg-MOLHIV} It is a molecular property prediction dataset adopted from the MoleculeNet, and among the largest of the MoleculeNet datasets, used for predicting the target molecular properties.
       
       \par 

    \section{the proofs of the theorems}
\subsection{proof of Lemma 3.4}\label{p:3.4}
\begin{minipage}[t]{0.62\textwidth}  %
\raggedright  %
\begin{proof}
Given a path containing a chord, we focus on the decomposition of the subpath that includes the chord, as the remaining subpaths are already chordless. In such cases, it suffices to select a single chord—preferably one whose endpoints are near the middle of the path—to partition it into chordless segments. This approach avoids exhaustive decomposition over all edges or chords, and is sufficient for our purposes.
\end{proof}
\end{minipage}
\hfill
\begin{minipage}[t]{0.35\textwidth}  %
\centering
\begin{tikzpicture}[scale=0.55, every node/.style={circle, draw, fill=white, inner sep=1.0pt}]
    \def\a{3}
    \def\b{1.5}

    \foreach \i in {1,...,7} {
        \pgfmathsetmacro{\angle}{180*(\i-1)/6}
        \pgfmathsetmacro{\x}{\a*cos(\angle)}
        \pgfmathsetmacro{\y}{-\b*sin(\angle)}
        \node (n\i) at (\x,\y) {\tiny \i};
    }

    \foreach \i in {1,...,6} {
        \pgfmathtruncatemacro{\j}{\i+1}
        \ifnum\i=2
            \draw[blue, thick] (n\i) -- (n\j);
        \else\ifnum\i=3
            \draw[blue, thick] (n\i) -- (n\j);
        \else\ifnum\i=4
            \draw[blue, thick] (n\i) -- (n\j);
        \else\ifnum\i=5
            \draw[blue, thick] (n\i) -- (n\j);
        \else
            \draw (n\i) -- (n\j);
        \fi\fi\fi\fi
    }

    \draw[red, thick] (n2) -- (n6)
      node[midway, above, draw=none, shape=rectangle, inner sep=1pt] {\footnotesize chord}
      node[midway, below=30pt, draw=none, shape=rectangle, inner sep=1pt, text=blue] {\footnotesize chordless paths};
\end{tikzpicture}
\end{minipage}
\subsection{proof of Lemma 4.3}\label{p:4.3}
\begin{figure}[H]
              \centering
              \includegraphics[width=0.45\textwidth, trim=1.3cm 5cm 4cm 3cm, clip]{
                     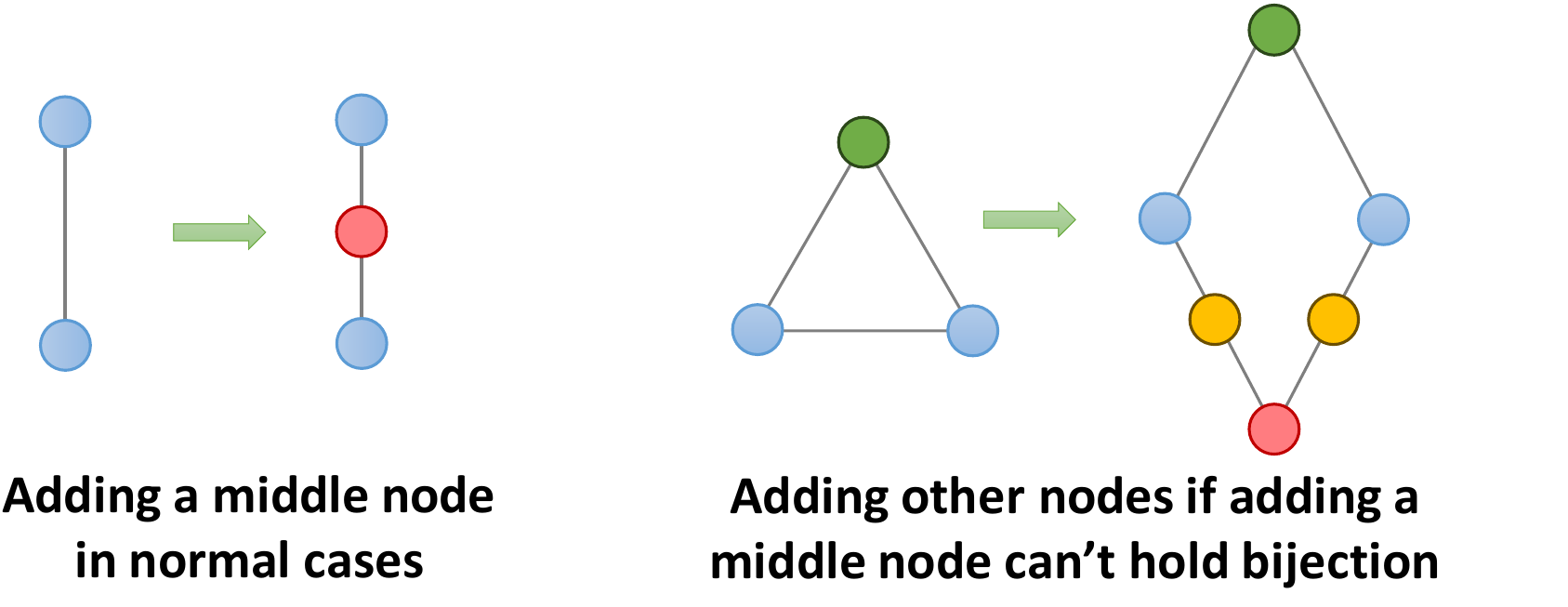
              }
              \caption*{Lemma 4.3: Construct $1-WL-colorable$ graphs by adding nodes.}
              \label{figure: for proof}
       \end{figure}

\begin{proof} After considering the $1-WL$ coloring, if there is
       an edge that has the same color as its neighbor, we can add a node in the
       middle of the edge, and then we can get a $1-WL-colorable$ graph. If the graph
       after adding the middle node has existed in the set, we can again add
       another two nodes in the middle of the edge, and so on (refer to Figure~\ref{figure: for proof}). By considering the
       finity of the set, we can get the $1-WL-colorable$ graphs in the set.
       
       \end{proof}
\subsection{proof of Theorem 4.9}\label{p:4.9}
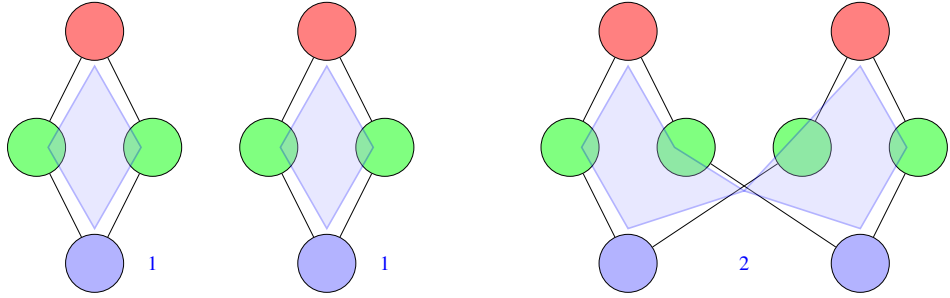
\begin{figure}[htbp]
\centering
\resizebox{0.9\linewidth}{!}{%
\begin{tikzpicture}[
    every node/.style={draw, circle, minimum size=1cm},
    rednode/.style={fill=red!50},
    greennode/.style={fill=green!50},
    bluenode/.style={fill=blue!30}
  ]

  \node[rednode] (R1) at (0,0) {};
  \node[rednode] (R2) at (4,0) {};

  \node[greennode] (G1) at (-1,-2) {};
  \node[greennode] (G2) at (1,-2) {};
  \node[greennode] (G3) at (3,-2) {};
  \node[greennode] (G4) at (5,-2) {};

  \node[bluenode] (B1) at (0,-4) {};
  \node[bluenode] (B2) at (4,-4) {};
  \node[draw=none,text=blue] at (1,-4) {1};
  \node[draw=none,text=blue] at (5,-4) {1};
  \draw (R1) -- (G1);
  \draw (R1) -- (G2);
  \draw (R2) -- (G3);
  \draw (R2) -- (G4);

  \draw (G1) -- (B1);
  \draw (G2) -- (B1);
  \draw (G3) -- (B2);
  \draw (G4) -- (B2);

  \draw[thick, draw=blue, fill=blue!30, opacity=0.3]
  (0,-0.6) -- (-0.8,-2) -- (0,-3.4) -- (0.8,-2) -- cycle;

\draw[thick, draw=blue, fill=blue!30, opacity=0.3]
  (4,-0.6) -- (3.2,-2) -- (4,-3.4) -- (4.8,-2) -- cycle;
\end{tikzpicture}
\hspace{2cm}
\begin{tikzpicture}[
    every node/.style={draw, circle, minimum size=1cm},
    rednode/.style={fill=red!50},
    greennode/.style={fill=green!50},
    bluenode/.style={fill=blue!30}
  ]

  \node[rednode] (R1) at (0,0) {};
  \node[rednode] (R2) at (4,0) {};

  \node[greennode] (G1) at (-1,-2) {};
  \node[greennode] (G2) at (1,-2) {};
  \node[greennode] (G3) at (3,-2) {};
  \node[greennode] (G4) at (5,-2) {};

  \node[bluenode] (B1) at (0,-4) {};
  \node[bluenode] (B2) at (4,-4) {};
  \node[draw=none,text=blue] at (2,-4) {2};

  \draw (R1) -- (G1);
  \draw (R1) -- (G2);
  \draw (R2) -- (G3);
  \draw (R2) -- (G4);

  \draw (G1) -- (B1);
  \draw (G2) -- (B2);
  \draw (G3) -- (B1);
  \draw (G4) -- (B2);

  \draw[thick, draw=blue, fill=blue!30, opacity=0.3]
  (0,-0.6) -- (-0.8,-2) -- (0,-3.4) -- (2,-2.75) -- (0.8,-2) --cycle;
  \draw[thick, draw=blue, fill=blue!30, opacity=0.3]
  (2,-2.75) -- (4,-3.4) -- (4.8,-2) -- (4,-0.6) -- cycle;

\end{tikzpicture}
}
\caption{Two WL-equivalent graphs that have different number of connected components}
\label{fig11}
\end{figure}
\begin{proof}
After removing the nodes in the last layer, we can get two trees that are 1-WL-equivalent, so they are isomorphic. So we pay attention to the nodes in the last layer to see their connection to the second-last layer.
Note that if there exists cycles that including each node in the last layer in each connected components, then we can distinguish the graphs with different number of nodes in the last layer in each connected components (As shown in the figure  \ref{fig11}, double $1$ vs a single $2$).\par
Otherwise, that is the case that  there exists a unique way between two nodes in the last layer since two paths help build a cycle.            
And then we can simplify the form of the graph to a three-layer form, since the equivalence of WL-tree above. 
We consider the color-equivalent mapping $\phi$ of a pair of nodes $u_{1},u_{2}$ in the last layer of the same WL-tree. And we denote the neighborhood of $u^{*}$ in another WL-tree by $u_{3}$. 
\item (1) if $\phi(u^{*})=u^{*}$ then the nodes in the last layer of the same WL-tree are connected by $u^{*}$ by color-equivalence. 
\item (2) if $\phi(u^{*})=u^{'}\neq u^{*}$ then $u_{1}$ and $\phi(u_{3})$ are neighbors of $u^{'}$ then we can find two paths.
\end{proof}

\end{document}